\documentclass[conference]{IEEEtran}
\IEEEoverridecommandlockouts
\usepackage{xtab}
\usepackage{cite}
\usepackage{amsmath,amssymb,amsfonts}
\usepackage{algorithmic}
\usepackage{graphicx}
\usepackage{array}
\usepackage{textcomp}
\usepackage{xcolor}
\usepackage{url}
\usepackage{xurl}
\usepackage{ltablex}
\usepackage{multirow}
\usepackage{fixltx2e}
\usepackage[numbers]{natbib}

\usepackage{lipsum}
\usepackage{makecell} 
\usepackage{color,array}
\usepackage{algorithmic}
\usepackage{graphicx}
\usepackage{natbib} %
\usepackage{pdflscape}
\usepackage{afterpage}
\usepackage{longtable}
\usepackage{colortbl}
\usepackage{longtable}
\newcolumntype{Y}{>{\centering\arraybackslash}m{3cm}}
\newcolumntype{M}{>{\centering\arraybackslash}m{1.8cm}}
\newcolumntype{C}{>{\centering\arraybackslash}m{1cm}}
\begin{document}

\title{Smart City Transportation: Deep Learning Ensemble Approach for Traffic Accident Detection\\
}
\author{\IEEEauthorblockN{Victor Adewopo}
\IEEEauthorblockA{\textit{School of Information Technology} \\
\textit{University of Cincinnati}\\
Cincinnati, USA \\
Adewopva@mail.uc.edu}
\and
\IEEEauthorblockN{Nelly Elsayed}
\IEEEauthorblockA{\textit{School of Information Technology} \\
\textit{University of Cincinnati}\\
Cincinnati, USA \\
elsayeny@ucmail.uc.edu}
}

\maketitle

\begin{abstract}
The dynamic and unpredictable nature of road traffic necessitates effective accident detection methods for enhancing safety and streamlining traffic management in smart cities. This paper offers a comprehensive exploration study of prevailing accident detection techniques, shedding light on the nuances of other state-of-the-art methodologies while providing a detailed overview of distinct traffic accident types like rear-end collisions, T-bone collisions, and frontal impact accidents. Our novel approach introduces the I3D-CONVLSTM2D model architecture, a lightweight solution tailored explicitly for accident detection in smart city traffic surveillance systems by integrating RGB frames with optical flow information. Our experimental study's empirical analysis underscores our approach's efficacy, with the I3D-CONVLSTM2D RGB + Optical-Flow (Trainable) model outperforming its counterparts, achieving an impressive 87\% Mean Average Precision (MAP). Our findings further elaborate on the challenges posed by data imbalances, particularly when working with a limited number of datasets, road structures, and traffic scenarios. Ultimately, our research illuminates the path towards a sophisticated vision-based accident detection system primed for real-time integration into edge IoT devices within smart urban infrastructures.
\end{abstract}


\section{Introduction}
The interconnection of road networks is one of the most challenging aspects of traffic flow prediction. Undoubtedly, a traffic incident at a major intersection can affect traffic flow and understanding of the spatial-temporal dynamics in road structures leading to accidents is becoming increasingly complex \cite{Wang2020a}. 
Accident detection and traffic analysis are critical components of smart city and autonomous transportation systems that can reduce accident frequency, severity, and improve overall traffic management. Road traffic accidents are a significant public health concern, accounting for 1.35 million fatalities and 50 million non-fatal injuries globally each year \cite{adewopo2023ai}.
There is a growing demand for intelligent transportation systems that can detect and track objects (vehicles, motorcycles, trains, buses, etc.) \cite{adewopo2022review}. Object detection in images has achieved significant performance in detecting and isolating objects in individual frames. However, in video data, which has gained high traction across different application areas, most image-based object detectors need to improve accuracy based on the spatio-temporal information available in the video dataset. Prior studies have widely explored temporal information for feature extraction in vehicle detection \cite{zhu2018towards}, while some other effective techniques utilize temporal information in post-processing. 
Recurrent neural networks (RNNs) cannot be used to extract spatial information from traffic data as they do not allow traffic sequences from different roads to be treated separately because of the nature of their implementation. Graph neural networks, on the other hand, combine the sequential data from time series data and graph neural networks together to provide a neural network that can be used for spatial and temporal data.
Le et al.~\cite{yu2021deep} study concluded that accident prediction at the road level is a crucial part of predicting accidents since accidents typically occur on a variety of roads and are affected by both internal factors (environment, road type, structure of the road) and external factors (such as the behavior of drivers, the weather, and the amount of traffic on the road). 
There is a clear distinction between traffic accident detection and traffic anomaly detection. Traffic anomaly encompasses a broader temporal window, including irregular movements without collisions, while accident detection is focused on a shorter window of traffic accidents defined by occurrences of car crashes that can be a subset of traffic anomalies \cite{fang2023vision}.
The perspective from which an accident is viewed can significantly influence the interpretation and subsequent analysis of the event. This study specifically targets accidents captured through traffic surveillance and dash cameras, concentrating on incidents involving collisions between various vehicle types or without colliding with other vehicles, excluding motorcycles. The diversity of accident scenes, combined with the multifaceted nature of the captured viewpoints, underscores the complexities inherent in accident detection, further exacerbated by factors such as harsh environmental conditions and the evolving nature of accident scenes.
The primary contributions of our research can be summarized as follows:
\begin{itemize}

\item We introduce an advanced vision-based accident detection system specifically optimized for real-time implementation on edge IoT devices such as Raspberry Pi. This minimizes computational overhead, making it ideal for smart city infrastructures and traffic surveillance systems.

\item We propose a  novel model architecture extracting RGB frames and optical flow information from video sequences, leveraging transfer learning techniques and the CONVLSTM2D structure to achieve improved performance accident detection, making our approach distinct from other prevalent methods.

\item Curated specialized benchmark dataset designed explicitly for accident detection from a traffic camera and dashcam point of view, encapsulating diverse roadway designs and scenarios, providing a rich resource for further research and development.
\end{itemize}

\section{Current Accident Detection Techniques}
\subsection{Deep Spatio-Temporal Convolutional Network (DSTGCN)}
In urban traffic management, accident detection systems help improve the quality of human lives in addition to providing adequate information for drivers regarding alternative routes. The study by Le et al. \cite{yu2021deep} indicates that traffic accidents are one of the most severe causes of fatalities and financial losses in our society at present. Thus, their work presents the Deep Spatio-Temporal Convolutional Network (DSTGCN) for predicting traffic accidents based on deep spatial and temporal data. Graph Neural Networks are a new domain that has shown remarkable improvement in capturing non-euclidean underlying graph structures. The framework proposed has three layouts:
\begin{itemize}
    \item \textit{Spatial layer learning:} Spatial data was captured from the road structures by identifying 20 points of interest (POI) that could capture the characteristics of the road structure. Based on the relative proximity of the identified point of interest, the probability of accident risk was calculated. Road-related structures are based on a cumulative calculation of points, road segments, and lengths of the road that have been determined by $\chi_{ij}^{P}=\left| \left\{ p:p\in \mathcal{D}^{pj}\wedge dist(\nu_i,p) \leqslant d \right\} \right|, for 1\leqslant j\leqslant 20$; where Xpi is denoted by the distribution of POIs around the road segment, and where $\mathcal{D}^{pj}$ represents the set containing POIs. $\nu_i,\left| .\right|$ represents road segments, $ dist(\nu_i,p)$ represents the distance between the road segment $\nu_i$ and POI.

\item \textit{Spatial-temporal learning:}
Based on the robust data collected, the researcher partitioned the road network into several grids and calculated the average speed within each of the grids. Using historical data from each road segment and an average traffic speed in each grid, the authors were able to extract the temporary traffic accident risk and represented it as $\chi_{i}^{temporal} =\left\{ \chi_{i}^{\nu, t} \right\}_{t=1}^{T}$.

\item \textit{Embedding layer for semantics and external representations:}
In the proposed framework, external factors that could influence accident risk were included when capturing the semantics as well as the external representations of the concepts. For all the road structures, the meteorological data collected is considered universal. The main external features that are taken into account are divided into two main categories: calendar features (day, month, timestamp) and meteorological features (weather type, temperature, humidity, wind speed, and direction) represented as $\chi^{external} =\left[\chi^{M,T} ; \chi^{C,T+1}  \right]$.
\end{itemize}

The proposed framework combines road network, point of interest (POI), taxi GPS, meteorological data, and traffic accident data together to extract the features and then incorporate them into a spatial convolution layer, spatial-temporal convolution layer, and embedding layer into a fully connected layer that learns the correlations between the different features in accident prediction. Compared to other state-of-the-art methods, such as stack denoising autoencoders (SdAE) and traffic accident prediction methods based on LSTMs (TARPML), the proposed method scaled higher. The decision tree algorithm showed a more impressive performance compared with other classical machine learning models in selecting relevant accident features and sensitivity to noise. The DSTGCN proposed model had higher precision, recall, and F1 scores than any other model, which is attributable to the robust information extracted from the three layers in the model architecture.

\subsection{Spatial Temporal Graph Neural Network (STGNN)}
The research of Wang et al.~\cite{Wang2020a} identified the dynamic influence of road networks, which cannot be accurately modeled by merely taking a look at their proximity. For learning long-range global dependencies in traffic flow, the researchers developed spatial-temporal graph neural networks. The proposed STGNN consists of three layers: A positional graph neural network, a recurrent neural network, and a transformer layer.
The architecture is based on the assumption that traffic $\mathcal{G}= \left ({v},\epsilon\right ), where~{v}= \left \{ \upsilon_{1},....,\upsilon_{N} \right \}$ is a set of ${N}$ traffic sensor nodes, where $\epsilon$ is a set of edges connecting the nodes.
Historically, traffic information is represented as  $\mathcal{Y}=\left (Y_{1},....,Y_{\mathcal{T}} \right)$ and $Y_{t} ~\epsilon ~\mathbb{R}^{N x 1}$ is the traffic information of nodes $\mathcal{N}$ at time ${t}$. The traffic flow prediction is based on modeling input $\mathcal{X} = \left( X_{1},....,X_{\mathcal{T}} \right)$ ~of length ~$\mathcal{T}$ ~to predict traffic in time ~$\mathcal{T}{}'$~\cite{Wang2020a}.
This model was trained with time $\mathcal{T}{}'$ by training STGNN with sliding historical data $Y_{t} + \mathcal{T}{}'$~\cite{Wang2020a}.
Spatial graph neural network (S-GNN) layers capture spatial-temporal information, whereas the GRU layer captures sequential temporal relations, while the transformer layer computes the attention for all positions at the same time based on the queries.
The framework proposed has a significant improvement over the baseline method when compared to predicting traffic flow across 15, 30, and 60 minute intervals with a MAPE of 15.6\%, 14.1\%, and 19.3\%, respectively.

\subsection{Single Shot MultiBox Detector (SSD)}
Taking into consideration the drawbacks of object detector algorithms in single frames that are highlighted in the research of Wang et al.~\cite{Wang2020a}, Yang et al.~\cite{Yang2021} attempts to improve object detection models through the introduction of a novel object detection algorithm developed specifically to handle video data and mitigate the biases of traditional models. This new model can handle object tracking and helps eliminate the limitations of single object detectors. The proposed feature fused Single Shot MultiBox Detector (SSD) can enhance the accuracy of object detection. This detector consists of two stages, namely:
\begin{itemize}
    \item Feature Fused SSD: The feature fused SSD object detection is based on the VGG16 network and other feature layers. Based on the drawback of SSD in detecting smaller objects, the feature maps of FC7 and CONV-2 are fused to capture semantic information and detect smaller vehicles. The detection is represented as $\begin{Bmatrix}
B_{t}^{B}=[b_{t1}^{B}, b_{t2}^{B},...,b_{tn}^{B}]
\\ a
S_{t}^{B}=[s_{t1}^{B}, s_{t2}^{B},...,s_{tn}^{B}]
\end{Bmatrix}$ ~\cite{Yang2021}.
    \item Tracking‑guided detections optimizing (TDO): The second phase of the algorithm is primarily dedicated to tracking the bounding boxes identified from frame {t} to frame {n} after the redundant bounding boxes have been removed by using a non-maximum suppression algorithm. A new vehicle is added to the subsequent frames when the distance between the overlapped boxes is greater than the chosen threshold, which is represented as $IoU(b_{ti}^{B},b_{tj}^{T}) <= threshold$. SSD models have comparative results with the Faster R-CNN. However, in terms of computational complexity and labeled data required, the SSD outperforms Faster R-CNN.
\end{itemize}
The model was trained on two popular datasets - ImageNet VIDs and Highway vehicle datasets. Due to data imbalance, the model appeared to be able to detect cars more than buses based on the higher number of cars in the train dataset. The results of the proposed model showed that the average precision for feature-fused SSD is 70.5 \% as compared to other state-of-the-art Faster R-CNN 63\%, SSD 67.5\%, and Tubelet Proposal Network (TPN) 68.4\%.

\subsection{Spatial-Temporal Mixed Attention Graph-based Convolution model (STMAG)}
In most cases, accidents are the result of unsafe human behaviors such as parking in an undesignated area or driving against the flow of traffic, etc. Existing accident prevention systems, like collision alarm systems, are useful to prevent accidents, but the systems tend to lack real-time alert systems that can trigger a warning signal in the event of an accident. A Spatial-Temporal Mixed Attention Graph-based Convolution model (STMAG) was proposed by Xiaoyang et al.  \cite{Wang2020} that relies on CNN and graph convolution networks to predict future accidents based on a spatial-temporal mixed attention graph. The method involves extracting spatial heterogeneity information from traffic data.
For the purpose of predicting accident at time ${t}$, the spatiotemporal relationship is modeled by learning a nonlinear relationship between the function $\mathcal{F}$ on the topology of the network ${G}$, and the multivariable input sequence $\mathcal{X}$, which is represented as $\hat{Y}_{T+j}=f\left( G;\left( X_{T-n},....,X_{T-1},X_T \right) \right)$. In this case, ${G}$ is the topological structure of the nodes; ${G}=(V,E)$ containing a set of nodes, ${X_T}$ is the concatenation of the traffic feature sets of the N nodes at the time of the target sequence. The main contribution of this research is to establish a mixed attention mechanism that can calculate the spatial and temporal dependencies between each node to predict incidents when anomalies are detected. Using the attention mechanism, the authors developed a mixed temporal and variable attention model for selecting information that correlates strongly with the target output. The temporal dependencies consist of an encoder, decoder, and intermediate state vector. Historical temporal dependencies were obtained in the encoder using a Gated Recurrent Unit with RNN. The output variable of the mixed attention has three categories of Safety, mildly dangerous, and severely dangerous to trigger early warning for a high probability of traffic accidents.
\subsection{Detection Transformer (DETR)}
Split seconds can make a significant difference between the occurrence of an accident and timely intervention. There have been many cases of people living with permanent disabilities as a result of the mismanagement of post-accident crises. In the study of Srinivasan et all.~\cite{Srinivasan2020}, the authors present a Detection Transformer and Random Forest classifier for the detection of accidents using surveillance cameras. The major contribution of the research is to reduce the complexity of accident detection and speed up the inference time for detecting road accidents. YOLO, Fast R-CNN, and Faster R-CNN are successful object detection algorithms but require annotating the train data and are computationally expensive. DETR, an algorithm from Facebook, was used to detect and track objects in conjunction with a Random Forest Classifier to detect accidents after they occurred. There are four layers in the proposed framework, including DETR architecture with CNN backbone, encoder-decoder blocks, fully connected layers, and a classification algorithm.
The input image ${x}$ learns the residual mapping $F(x)=H(x)-x$. The number of decision trees chosen for the classification layer is 500, and the tree depth is set at 40. The entropy is found to produce improved results and is calculated by $Entropy= \Sigma_{i=1}^{N} \left( -p_{i} \times log(p_{i}) \right)$. The proposed framework had a detection rate of 78\% but lags behind in the work presented by wang et al.~\cite{wang2020vision}.

\subsection{Scale-invariant Feature Transform (SIFT) and Sparse Topic Model}
The Xia et al.~\cite{Xia2018} paper presents an interesting approach to treating videos as documents and trajectories as topics using the scale-invariant feature transform (SIFT) and sparse topic model. The researchers proposed a method for detecting anomalies in traffic surveillance. A deviation from normal traffic flow could indicate an anomaly in the patterns of motion. Several motion pattern analyses have been performed on video streams in order to identify individual anomalies. The slight difference between normal and abnormal motion patterns makes it difficult to detect abnormal traffic events. The authors presented a sparse topic model based on a probability density function expressed in a non-probabilistic form to describe each trajectory in the video based on the Fisher kernel method. Compared to optical flow methods, SIFT was used to track the trajectory after features were extracted using the dense trajectory method based on its more robust performance.

A sparse topic model was created by taking the video as a document and assuming there are ${\mathrm{K}}$ topics present in the video as a topic dictionary represented as matrix $\mathrm{D} \in \mathrm{R}^{K \times N}$. The whole document is represented as the code set $\alpha_\mathcal{d}=(\alpha_\mathrm{d,1}, \alpha_\mathrm{d,2}, ...,\alpha_\mathrm{d,N})^T$ and the dictionary ${\mathcal{D}}$. Considering that the model was trained on different videos that may differ from the dataset for detecting abnormalities, a constraint term of $\ell_1$-norm was applied to the column vector ${\mathcal{D}}$. Based on the developed model, new evaluation dataset topics are generated by calculating the proportion of each topic to the document ${d}$ on the ${k-th}$ topic as: $ \Theta_{d,k} =\frac{{\underset{{n\in I_d}}{\sum}{\alpha_{d,nk} D_{kn}}}}{{\underset{{n\in I_d}}{\sum} {\underset{{k\in K}}{\sum}}{\alpha_{d,nk} D_{kn}}}}$. The similarity between the two clips is calculated by taking the distance between clip $\mathrm{d_i}$ and $\mathrm{d_j}$ using $dis(\mathrm{d_i,d_j})=-lg\left({\overset{}{\sum_{k=1}^{k}}}{\sqrt{\Theta_{d_i,k}{\Theta_{d_j,k}}}}\right)$. 
The number of word/clip documents appearing in the test clip must exceed a threshold in order to flag the clip as anomalous.

The framework is designed to scan every topic (trajectories) that appears in the document (video) and classify that video as anomalous if the topic number exceeds a certain threshold. The proposed method was evaluated on the QMUl and AVSS datasets. Compared to other motion analysis methods, the method is effective with an AUC score of 91.2\% on the AVSS dataset (JSM is based on motion trajectory and has an AUC score of 80.2\%, while STC and GPR are representative video analysis methods for detecting anomalies and their AUC scores are 85 and 84\% respectively). 
\subsection{Mask R-CNN}
Similar to Xia et al.~\cite{Xia2018} study using a detection algorithm. Ijjina et al.~\cite{Ijjina2019} present a framework for detecting car accidents from traffic surveillance by training deep learning models (Mask R-CNN) to detect anomalies in car trajectory or speed after an overlap. The authors aim to identify traffic accidents spontaneously, reducing human bias and the time taken to effectively identify and respond to accidents. Through an automated system for contacting the traffic department and paramedics in cases of accidents, intelligent systems can potentially scale out human performance.
Mask R-CNN is an improved state-of-the-art methodology for segmenting images and tracking multiple objects in a single image, which is a difficult task for CNN. Multiple objects can be tracked in an image by using region-based CNN (R-CNN) and fast region-based CNNs. Mask R-CNN offers a number of advantages over Faster R-CNN, including the ability to separate objects from backgrounds in images and identify individual objects by means of semantic and instance segmentation. Additionally, it is a faster network despite being able to localize, segment, and classify objects.
Ijjina et al.~\cite{Ijjina2019} present a framework that can be adapted to any CCTV camera and can be easily scalable. The three-phased approach involves detecting the object (vehicle), tracking the movement of the vehicle, and finally, detecting the accident based on the extracted features. The object detection phase utilizes the Mask R-CNN to generate the bounding boxes, class IDs, and masks for objects in an image. Mask R-CNN uses the same architecture as Faster R-CNN with the addition of masks to objects detected in the feature map. In the second step, the objective is to identify unique vehicles and objects across a large set of frames (object tracking and feature extraction). Using a centroid tracking algorithm, vehicles are tracked and recorded in subsequent frames by recording their coordinates and Euclidean distances between centroids. A unique number is assigned to each object appearing in subsequent frames, and an object is removed once it is no longer visible. Based on the Euclidean distance, trajectory change, intersection angle, and speed of an object, predictions are made at the final stage. Based on a set threshold, the acceleration anomaly, trajectory anomaly, and angle anomaly are read for triggering an accident anomaly.
The system proposed by Ijjina et al.~\cite{Ijjina2019} could significantly reduce the number of false positives for cars traveling from opposite directions or lanes. However, the methodology relies on tracking anomalies after vehicles overlap, which makes identifying accidents in the same direction difficult. A similar study by Singh et al.~\cite{singh2018deep} sought to determine the probability of accidents occurring by tracking vehicles in opposite directions using spatial-temporal video volume information. There is no explanation of how the threshold for detecting anomalies was selected or calculated by the authors. The model was evaluated under different weather conditions, including daylight, low feasibility, rain, hail, and so on. Comparing the proposed framework with other existing models, the proposed framework achieved a 71\% detection rate and a false alarm rate of 53\%. The deep Spatio temporal model performed better with a 77.5\% detection rate.

\section{Traffic Accidents}
\begin{figure}[htbp]
\centerline{\includegraphics[width=0.9\linewidth]{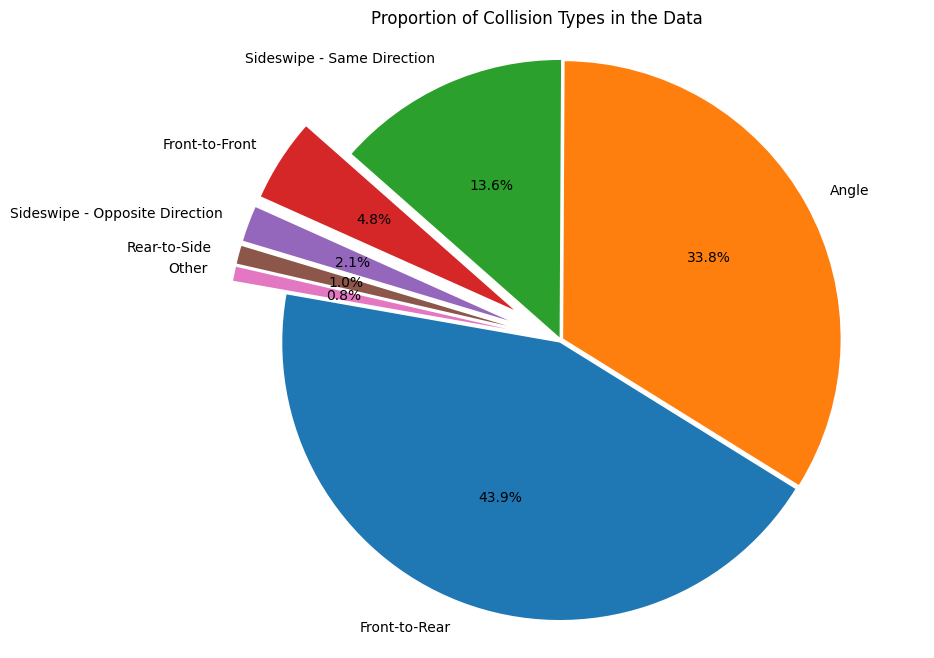}}
\caption{Percentage distribution of different types of collision.} 
\label{collision}
\end{figure}

Traffic accidents are a global issue that causes many injuries, fatalities, and property damage every year~\cite{mohammed2019review}. In order to reduce the negative impact of traffic accidents, it is essential to understand the various types of accidents and their causes. Among the many types of traffic accidents, rear-end collisions, T-bone or side impact collisions, and front impact collisions are the most common and pose a significant threat to road safety.
Our study will discuss only the main types of accidents: rear-end collisions, T-bone or side impact collisions, and front-impact collisions. Figure \ref{collision} showcases the percentage distribution of collision types with Fron-to-Rear collision accounting for 43.9\% of traffic accidents, 33.8\% accounts for angle, and 13.6\% accounting for same-direction traffic accidents. For effective interventions to reduce their frequency and severity and to develop effective strategies that can reduce their occurrence and severity and improve road safety, it is essential to understand the factors contributing to these accidents.

\subsection{Rear-end collision} 
According to the National Highway Traffic Safety Administrator, Rear-end collisions are the most common type of traffic accident, accounting for 29\% of all crashes and causing significant injuries and fatalities each year \cite{Hs2007AnalysesOR}. This type of accident occurs when one vehicle collides with the back of another vehicle traveling in the same direction. Rear-end collisions can result in injuries, from minor whiplash to severe head and spinal injuries and even fatalities. Several factors contribute to rear-end collisions, including distracted driving, following too closely, sudden stops, and poor weather conditions. One of the leading causes of rear-end collisions is driver distraction, such as texting or talking on the phone while driving. Additionally, distracted drivers cannot react quickly enough to avoid a collision in the event of sudden stops or slow down of vehicles ahead \cite{eboli2020factors}. 
Most of these accidents involve a stopped or slow-moving lead vehicle. Efforts to address the issue have had limited success, with the center high-mounted stop lamp (CHMSL) being one of the most notable initiatives launched in 1986. Although the CHMSL has reduced rear-end collisions by 4\%, further improvement is still needed. In 1999, the National Highway Traffic Safety Administration (NHTSA) contracted with the Virginia Tech Transportation Institute (VTTI) to conduct tests and make recommendations for improved rear lighting and signaling systems \cite{Hs2007AnalysesOR}. To prevent rear-end collisions, drivers can manually employ safety measures such as increasing the distance between vehicles, reducing speed in poor weather conditions, and avoiding distractions while driving. Advanced driver assistance systems (ADAS) can be installed in vehicles to help drivers avoid collisions. These systems use sensors and cameras to detect potential collisions and warn the driver \cite{kukkala2018advanced}. Developing new intelligent transportation systems (ITS) can help prevent rear-end collisions. These systems use connected vehicle technology to allow vehicles to communicate with each other (V2X communication) and with infrastructure (Collision avoidance, lane detection), providing drivers with real-time information about road conditions and potential hazards (indoor monitoring). For example, if a vehicle ahead suddenly brakes, the system can alert other nearby vehicles to take appropriate action, such as slowing down or changing lanes \cite{kukkala2018advanced}.

\subsection{T-Bone Collision}
T-bone accidents are also often referred to as side-impact collisions. T-bone accidents are likely to be severe injuries based on the lack of structural barrier between driver and passengers engaged in such events, depending on the advanced safety features available in the car. This type of collision occurs when the front end of one vehicle strikes the side of another vehicle at a perpendicular angle. This type of collision is hazardous because the sides of vehicles have less protection compared to the front and rear. T-bone collisions can result in serious injuries or fatalities for occupants of the struck vehicle \cite{mohammed2019review}.
According to the National Highway Traffic Safety Administration (NHTSA), T-bone collisions account for approximately 13\% of all passenger vehicle occupant fatalities in the United States. In addition, side impact collisions are the second most common type of fatal collision for passenger vehicle occupants, after frontal crashes \cite{nhtsa}. 
Technological advancements have also been developed to mitigate the severity of side-impact collisions. For example, side airbags are now standard in many new vehicles and can provide additional protection to occupants in a side-impact collision. Moreover, some vehicles are equipped with advanced safety features such as automatic emergency braking and lane departure warning systems, which can help prevent collisions from occurring in the first place.
Side impact collisions are a serious safety concern on the roads \cite{kukkala2018advanced}. Efforts to improve safety through engineering and technology continue to be essential to prevent or mitigate the severity of T-bone collisions. The research of Eboli et al. \cite{eboli2020factors}, found some correlation between road structures, driver factors, and types of accident collision through the analysis of road accident type. Surface conditions, such as dry road surfaces, reduce the probability of a front/side collision in serious accidents. Driver-related factors, such as having a car license, increase the probability of a front/side collision in serious accidents. Furthermore, environmental factors, such as sunny weather, increase the probability of a front/side collision. The authors also pointed out that driver-related factors play a more important role in the probability of a front/side collision \cite{eboli2020factors}.

\subsection{Frontal impact accident}
Frontal impact accident accounts for the most inter-vehicular accident with severe injuries and deaths; it is the frontal impact with 35\% severity impact \cite{jirovsky2015classification}. Front-end collision occurs when the front end of one vehicle collides with the front end of another vehicle or when a vehicle collides with a fixed object, such as a wall or tree. Figure \ref{collision} illustrates that only 4.8\% of traffic accidents are due to frontal collisions. These collisions often result in serious injuries or fatalities due to the force of impact. Front-end collisions can occur for a variety of reasons, including driver error, speeding, distracted driving, and poor weather or road conditions. In some cases, faulty vehicle equipment or manufacturing defects may also contribute to front-end collisions. Front-end collisions can be classified into two main categories: offset and full-frontal collisions. In an offset collision, the front of one vehicle collides with the side of another vehicle, resulting in a twisting motion that can cause significant damage to both vehicles. Full-frontal collisions occur when the entire front end of one vehicle collides with the front end of another vehicle, resulting in a direct impact that can cause severe injuries or death \cite{nolan2001frontal}.
Efforts have been made to reduce the incidence and severity of front-end collisions. This includes using safety features such as airbags, crumple zones, and seat belts, as well as developing advanced driver assistance systems (ADAS) that can alert drivers to potential collisions and even take action to avoid them. The existing forward collision warning (FCW) systems based on kinematic or perceptual parameters have some drawbacks in warning performance due to poor adaptability and ineffectiveness \cite{sheoreysensor}. To address these problems, machine learning and deep learning algorithms have been proposed. However, these models lack consideration for multi-staged warnings, which could distract or startle the driver. A light gradient boosting machine (LGBM) learning algorithm was used to develop a multi-staged FCW model, which was evaluated using a driving simulator by twenty drivers \cite{ma2022adaptive}. The study found that the front vehicle acceleration, time-to-collision (TTC), and relative speed strongly affected the warning stages from the proposed model. The authors aim to develop LGBM for developing FCW models that could improve warning performance while considering multi-staged warnings \cite{ma2022adaptive}.
The study of Jirovsky et al. \cite{jirovsky2015classification} employed a new approach for determining the probability of collision between two vehicles by defining a 2-D reaction space, which describes all possible positions of the vehicles in the future. This approach enables mitigation of collision by exploring alternative causes of action such as changing direction in addition to braking \cite{jirovsky2015classification}.
\begin{figure}[htbp]
\centerline{\includegraphics[width=0.9\linewidth]{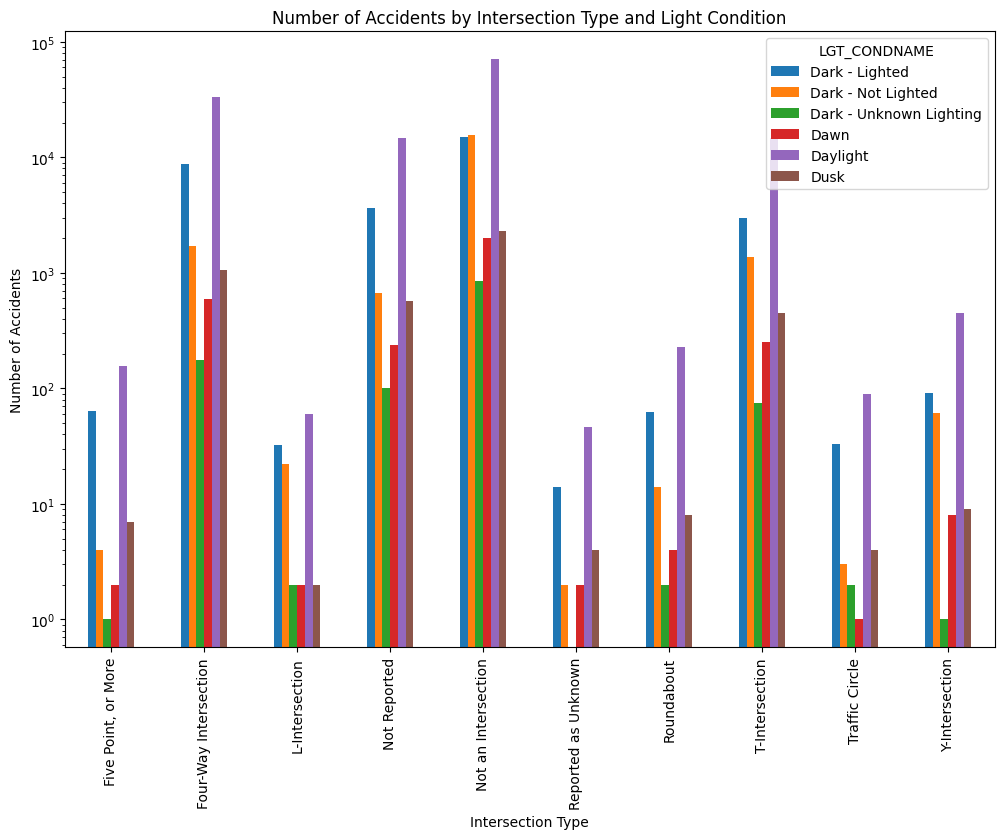}}
\caption{Accidents by Intersection Type and Light Condition Associated.} 
\label{intersection}
\end{figure}
Figure \ref{intersection} shows the analysis of traffic accidents at different intersections and the light conditions. Light conditions play a crucial role in traffic safety, and the data indicates that accidents are more common at intersections, with four-way intersections being hazardous. Whether lighted or not, the number of accidents in dark conditions is high across all intersections. The data also shows that accidents in dark-light conditions are more frequent at T-intersections and four-way intersections.
Adequate road lighting is crucial for ensuring the safety of drivers during nighttime driving. The study by Alharbi et al. \cite{alharbi2023performance} on the performance appraisal of urban lighting systems aligns with the findings in Figure \ref{intersection}. The researchers found that electronic billboard positioning, oncoming vehicle lights, and poor lighting conditions during inclement weather, particularly dust, are significant factors affecting the performance of urban street lighting systems (USLSs) as perceived by road users. The research by Bridger et al. \cite{bridger2012lighting} reported that modern LED lighting is a disruptive technology and that decreasing nighttime fatalities and injuries due to modern road lighting has significant cost benefits. On the other hand, the study by Marchant et al.~\cite{marchant2022determine} found limited evidence to support road safety improvement through relighting traffic lights with white lamps in the UK. This further highlights the importance of continued research and evaluation of road lighting systems and their impact on road safety.  

\section{Data Collection}
A comprehensive data collection process was implemented due to the lack of readily available robust datasets in this domain. Here, we provide a detailed description of our collection methodology and sources.
\subsection{Types of Data and Sources}
Two primary categories of data were collected:

\begin{itemize}
    \item  \textbf{\textit{Traffic/Surveillance Data (Trafficam):}} This data contains videos captured from traffic and surveillance cameras situated at various locations. Trafficam provides a unique vantage point for capturing vehicular movement. These cameras are strategically positioned at various locations to offer an aerial or elevated view of the road, enabling a holistic view of vehicles. This perspective is essential for several reasons, including;
    Trajectory Angle: Trafficam videos capture the trajectory angle of vehicles. This angle denotes the path and direction of a vehicle’s movement, which is valuable for understanding vehicular dynamics and causal factors leading to an accident.
    Full Car View: The elevated vantage point of traffic and surveillance cameras provides a full silhouette and profile of vehicles crucial for detecting anomalies or changes in vehicular posture, like tilting during a potential rollover.
    Datasets from the TrafCam angle are particularly beneficial for accident detection because the overhead view minimizes occlusions, allowing for an unobstructed view of potential accident sites, and monitoring multiple vehicles simultaneously can help detect and analyze multi-vehicle collisions.

    \item \textbf{\textit{Dash Camera Video (DashCam):}} These videos are typically recorded from cameras installed on vehicle dashboards. DashCam videos offer a ground-level, front-facing perspective from vehicles, capturing the road ahead and occasionally vehicle interior/rear view. While invaluable in many respects, DashCams are beneficial for providing firsthand accounts of incidents and useful for understanding drivers' accident viewpoints, including capturing close-up details of incidents. Some of the challenges of Dashcams are the restricted field of view (focus mainly on the road ahead), Camera view occlusion, and variability in video quality based on different brands and models.
\end{itemize}
In order to curate a robust, diverse dataset, we collected videos from a wide array of sources and different geographic locations worldwide. The majority of our video data originates from platforms like YouTube. We employed strategic keyword searches to identify and curate relevant videos. Keywords included terms like "traffic accident," "traffic camera accidents," and "car accidents." In order to maximize the geographic and linguistic diversity of our dataset, these keyword searches were conducted in multiple languages, including English, French, Spanish, and Russian.
Table [\ref{TrafficDataCount}] provides a detailed breakdown of traffic and dashcam videos, categorized by accident occurrences and normal traffic conditions. The data is further divided into training, validation, and testing sets. Table [\ref{TrafficDataCount}] showcases the distribution of these categories across different data types, including Trafficam, Dashcam, and External Data sources. In order to supplement our curated dataset. We utilized an external data source (Car Crash Dataset) publicly available on GitHub \cite{BaoMM2020}. We aggregated approximately 5,700 datasets to train and evaluate our model performance.

\begin{table*}[h]
	\centering
	\caption{Distribution of traffic and dashcam footage across different data types.}
    \label{TrafficDataCount}
    \begin{tabular}{|
    >{\columncolor[HTML]{FFFFFF}}p{1.2cm}|
    >{\columncolor[HTML]{FFFFFF}}p{1.2cm}|
    >{\columncolor[HTML]{FFFFFF}}p{1.2cm}|
    >{\columncolor[HTML]{FFFFFF}}p{1.2cm}|p{1.2cm}|p{1.2cm}|p{1.2cm}|p{1.2cm}|p{1.2cm}|p{1.2cm}|}
    \hline
   {\textbf{Data Type}} & {\textbf{Trafficam Accident}} & {\textbf{Trafficam Normal-Traffic}} & {\textbf{Total Trafficam Dataset}} &{\textbf{Dashcam Accident}} & {\textbf{Dashcam Normal-Traffic}} & {\textbf{Dashcam Total Dataset}} &{\textbf{Ext. Data Accident}} &{\textbf{Ext. Data Normal-Traffic}} & {\textbf{Total DataSet}} \\ \hline
    Train & 603 & 511 & 1114 & 763 & 639 & 1402 & 1250 & 146 & 3912 \\ \hline
    Val & 190 & 140 & 330 & 268 & 224 & 492 & 150 & 82 & 1054 \\ \hline
    Test & 134 & 60 & 194 & 196 & 154 & 350 & 100 & 81 & 725 \\ \hline
    \end{tabular}
\end{table*}

\subsection{Video Processing and Annotation}
To ensure our dataset is concise and relevant, videos were processed based on the rapid dynamics of accidents. We segmented the videos into five-second non-overlapping clips, ensuring that each segment captures a distinct event or scene. This segmentation strategy aids in minimizing redundancy and focusing on the most relevant content for accident recognition.
Each 5-second video segment was annotated manually using the Labelbox annotation tool. Annotations provide information about the type of accident depicted in the video. Categories included, but were not limited to, ``all over", ``front end," ``rear end," ``side hit," and ``normal traffic." Due to the limited dataset in each category, our final dataset was grouped into two distinct categories (Traffic Accident and Normal Traffic). The camera resolution impacts the clarity and detail of the captured video, with higher resolution offering finer details. In this study, regardless of the initial resolution, videos are processed and cropped to 224x224x3 to conform to the I3D model input requirements. This standardized resolution maintains consistency and ensures comparability in results. We also performed video normalization by dividing pixel values by 255 and scaling the data, aiding the model in faster and more stable convergence.
To manage computational limitations, we strategically decreased our model's input from 150 to 50 frames, incorporating every third frame and subsequently to 30 frames by utilizing every fifth frame. This consideration was crucial in ensuring the effective detection of swift, brief actions for identifying accidents. The optimized frame selection was calibrated to retain the integrity of rapid actions, balancing the need for computational efficiency and resource conservation.

\section{Methodology}
Machine learning algorithms are increasingly being used in accident detection. These algorithms can be trained on large datasets of accident data to detect and classify accidents based on various features such as speed, direction, and vehicle type. This method is becoming more popular due to its ability to adapt to changing road conditions and detect real-time accidents. Singh et al.~\cite{singh2018deep} proposed a framework that extracts deep representation using autoencoders and an unsupervised model (SVM) to detect the possibility of an accident.
A deep learning model using computer vision is another promising method for accident detection. In this method, cameras are used to capture real-time video footage of the road and surrounding environment. 
Zadobrischi et al. \cite{zadobrischi2022intelligent} focuses on the integration of traffic monitoring systems into intelligent transport systems (ITS) to properly manage traffic and reduce the negative impact of congestion and road accidents in real-time using video and image data. 
Chan et al.~\cite{chan2016anticipating} proposed a Dynamic-Spatial-Attention (DSA) Recurrent Neural Network (RNN) for anticipating accidents in dashcam videos based on the vehicle trajectory and motion. The developed algorithm contains an object detector to dynamically gather subtle cues and the temporal dependencies of all cues to predict accidents two seconds before they occur. Ghahremannezhad et al.~\cite{ghahremannezhad2022real} introduce a three-step hierarchical framework for detecting traffic accidents at intersections using surveillance cameras. A unique cost function is utilized during object tracking to handle occlusions, overlapping objects, and object shape changes.
\subsection{I3D-CONVLSTM2D Model Architecture for Accident Detection}
\begin{figure}[htbp]
\centerline{\includegraphics[width=0.85\linewidth]{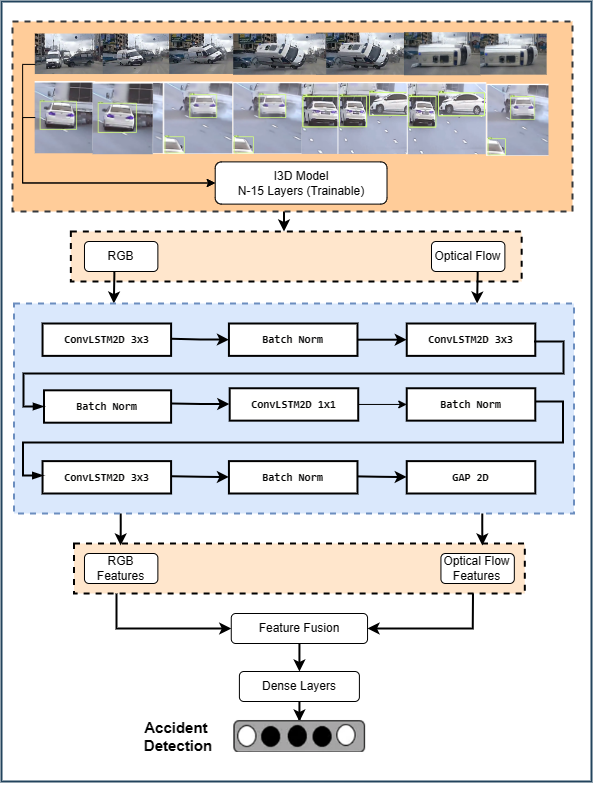}}
\caption{Accidents Detection Framework.} 
\label{Accident_architecture}
\end{figure}

The proposed architecture shown in Figure [\ref{Accident_architecture}] leverages the power of the Inflated 3D ConvNet (I3D) model, which is designed to capture spatiotemporal information from video frames. The architecture is divided into two main branches: RGB and Optical Flow (OF). Both branches are processed through the I3D model and further refined using ConvLSTM2D layers.
\begin{itemize}
    \item  \textbf{\textit{Input Video Frames:}} Given a video \( V \) with \( N \) frames, each frame \( f_i \) is of size \( 224 \times 224 \times 3 \) for RGB and \( 224 \times 224 \times 2 \) for Optical Flow.
\[V = \{f_1, f_2, \ldots, f_N\}\]
The video frames are divided into overlapping windows. Each window \( w_j \) contains a sequence of frames, where \( T \) is the temporal depth of the window.\[w_j = \{f_j, f_{j+1}, \ldots, f_{j+T}\}\]
\item  \textbf{\textit{I3D Model:}}
The Inflated 3D ConvNet (I3D) model was introduced in 2017 by Carreira et al.~\cite{Carreira2017} is a deep learning architecture designed for video recognition tasks. The I3D model is a 3D convolutional neural network that captures both spatial and temporal information. The model consists of multiple 3D convolutional layers, pooling layers, and inception modules. The last 15 layers of the I3D model are set to trainable to fine-tune the model for accident detection \cite{Carreira2017}.
\item  \textbf{\textit{ConvLSTM2D Layers:}}
RGB and Optical Flow branch of the I3D model is passed through three ConvLSTM2D layers. The ConvLSTM2D layer is a combination of convolutional and LSTM layers, designed to capture spatial and temporal dependencies in sequence data \cite{yadav2023human, adewopo2023baby, elsayed2018empirical}.
Given an input tensor \( X \) of shape \( T \times H \times W \times C \), the ConvLSTM2D layer computes~\cite{elsayed2020reduced}:
\begin{align*}
F_t &= \sigma(W_f \ast X_t + U_f \ast H_{t-1} + b_f) \\
I_t &= \sigma(W_i \ast X_t + U_i \ast H_{t-1} + b_i) \\
O_t &= \sigma(W_o \ast X_t + U_o \ast H_{t-1} + b_o) \\
C_t &= F_t \odot C_{t-1} + I_t \odot \tanh(W_c \ast X_t + U_c \ast H_{t-1} + b_c) \\
H_t &= O_t \odot \tanh(C_t)
\end{align*}
where: \( \sigma \) is the sigmoid activation function, \( \odot \) denotes element-wise multiplication and \( W \), \( U \), and \( b \) are the weights and biases of the ConvLSTM2D layer.

\item  \textbf{\textit{Global Average Pooling 2D (GAP2D):}}
The GAP2D layer computes the average value of each feature map, reducing the spatial dimensions while retaining the depth.
\[GAP(X) = \frac{1}{H \times W} \sum_{i=1}^{H} \sum_{j=1}^{W} X_{i,j} \]



\item  \textbf{\textit{Dense Layers and Prediction:}}
The combined feature vector is passed through multiple dense layers with ReLU activation. The final prediction layer uses a softmax activation to classify the video window as ``Accident" or ``No Accident". where \( z_1 \) and \( z_2 \) are the logits for the two classes.
\begin{align*}
P(\text{Accident}) &= \frac{e^{z_1}}{e^{z_1} + e^{z_2}} \\
P(\text{No Accident}) &= \frac{e^{z_2}}{e^{z_1} + e^{z_2}}
\end{align*}

The proposed architecture effectively captures spatiotemporal information from video frames and uses it to detect accidents. The combination of the I3D model, ConvLSTM2D layers, and feature fusion ensures that both spatial and temporal dependencies are considered, leading to accurate accident detection.

\end{itemize}

\section{Experimental Result}
\begin{table*}
\centering
\caption{Summary of Experimental Result}
\label{Results}
\footnotesize 
\renewcommand{\arraystretch}{1.3}
\begin{tabular*}{\textwidth}{@{\extracolsep{\fill}} MYYCCCccc}
\textbf{Model Name} & \textbf{Loss} & \textbf{Accuracy} & \textbf{Time}& \textbf{Precision} & \textbf{Recall} & \textbf{F1}& \textbf{Acc.} & \textbf{MAP} \\
\hline\hline
I3D-CONVLSTM2D RGB Only & \includegraphics[width=3cm,height=3cm]{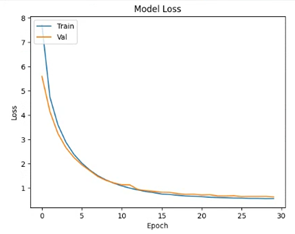} & \includegraphics[width=3cm,height=3cm]{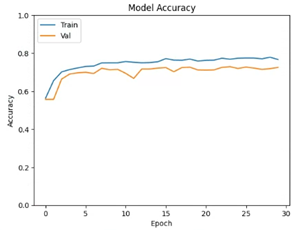} & 110 & 0.73 & 0.72 & 0.72 & 0.72& 0.78 \\
\hline
I3D-CONVLSTM2D Non-Trainable RGB + Optical flow & \includegraphics[width=3cm,height=3cm]{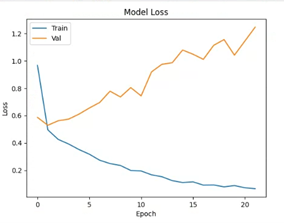} & \includegraphics[width=3cm,height=3cm]{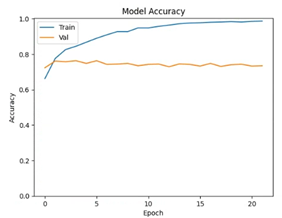} & 120 & 0.75 & 0.75 & 0.75 & 0.75& 0.81 \\
\hline
I3D-CONVLSTM2D Augmented RGB + Optical flow & \includegraphics[width=3cm,height=3cm]{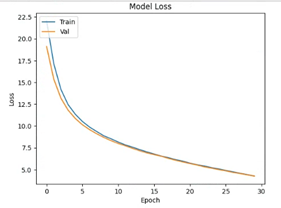} & \includegraphics[width=3cm,height=3cm]{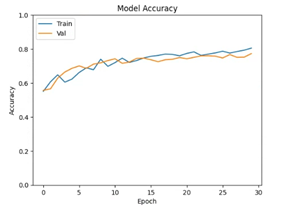} & 150 & 0.79 & 0.79 & 0.79 & 0.79& 0.86 \\
\hline
I3D-CONVLSTM2D Trainable RGB + Optical flow & \includegraphics[width=3cm,height=3cm]{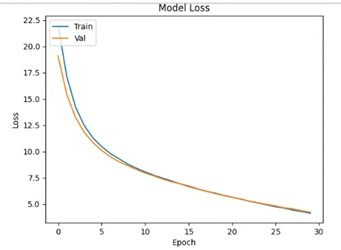} & \includegraphics[width=3cm,height=3cm]{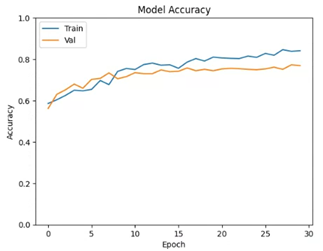} & 130 & 0.80 & 0.80 & 0.80 & 0.80 & 0.87\\
\hline
DenseNet-Transformer RGB Only~\cite{huang2017densely, vaswani2017attention} & \includegraphics[width=3cm,height=3cm]{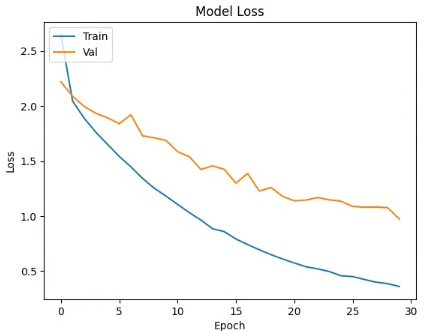} & \includegraphics[width=3cm,height=3cm]{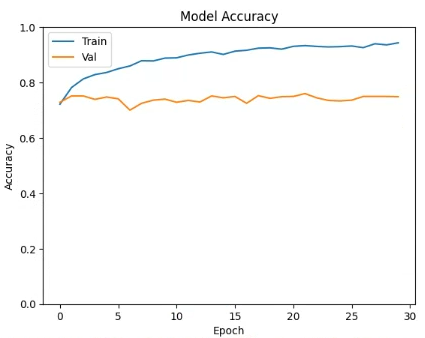} & 300 & 0.75 & 0.75 & 0.75 & 0.74 & 0.80\\
\hline
\end{tabular*}
\end{table*}

In this section, we provide a comprehensive analysis of the performance of four distinct models for action recognition within the context of accident detection, as shown in Table (\ref{Results}).
A recent study by Chen et al. \cite{chen2023hierarchical} introduced a novel Hierarchical Accident Recognition Method for Highway Traffic Systems that consists of Object detection, object tracking, car speed estimation using optical flow, and accident detection threshold based on proximity. Similarly, Zhu et al. \cite{zhu2023novel} proposed an alternative accident detection method using traffic flow features, road congestion, traffic intensity, and traffic state instability (flow and speed) to provide a more comprehensive representation of traffic conditions, enhancing the ability of the system to detect anomalies.
As detailed in this paper, our approach to accident detection stands out from existing research due to our emphasis on extracting RGB frames and optical flow information from video frame sequences. This is achieved through the utilization of transfer learning techniques and the CONVLSTM2D model architecture, as illustrated in Figure [\ref{Accident_architecture}]. To capture the distinct characteristics of accident motion effectively, we employ a ConvLSTM2D network to mitigate the limitations observed in basic CNN, RNN, and LSTM networks, resulting in a significant improvement in accident detection precision. Consequently, our approach represents a unique and valuable contribution to autonomous accident detection in smart cities. We explore how different architecture and parameter configurations are able to separate features indicative of accidents, leading to improved accident detection model performance.

\subsection{I3D-CONVLSTM2D RGB Only}

The I3D-CONVLSTM2D RGB Only model aimed to identify accidents by analyzing RGB frames exclusively in the input video. This model utilized a relatively modest configuration of ConvLSTM2D layers with 64, 32, and 32 filters. Additionally, dropouts were applied to mitigate overfitting.
Despite its simplicity, this model showcased good performance in distinguishing accident-related features. It achieved mean average precision (MAP) of 78\%, accuracy, precision, recall, and F1 score of 72\%, respectively, after training on 30 epochs on extracted features from the I3D Model, signaling its ability to capture some of the critical cues associated with accident recognition.

\subsection{I3D-CONVLSTM2D Non-Trainable RGB + Optical Flow}
The second model iteration adopted a nuanced approach to training by explicitly setting all layers of the I3D component to non-trainable, transforming it into a fixed feature extractor. We removed the classification head of the I3D model, replacing it with multiple layers of CONVLSTM2D designed to process both RGB frames and optical flow.
The model performance did not show notable improvement with a restrained learning rate set at \(1 \times 10^{-4}\), the accuracy, precision, recall, and F1 score on test data plateaued at a value of 75\% and MAP of 81\%. This model performance underscores the assertion that transitioning to a non-trainable I3D configuration, paired with CONVLSTM2D layers, can not succinctly capture relevant spatio-temporal features for accident detection using the I3D transfer-learning explicitly as a feature extractor.

\subsection{I3D-CONVLSTM2D Augmented RGB + Optical Flow}
The third model, I3D-CONVLSTM2D Augmented RGB + Optical Flow, explored the impact of data augmentation techniques on accident detection. Specifically, our augmentation strategy comprised cropping, zooming, and careful rotation. Notably, rotations were cautiously capped to ensure realistic representations, and vertical flipping was avoided to prevent generating videos that appeared upside-down. Video augmentation, including flipping and rotation, was applied to the dataset, aiming to augment temporal feature learning and enhance the recognition of accident-related features.
Contrary to our initial hypothesis, the integration of these augmentation techniques did not precipitate the anticipated enhancement in accident detection. The model had a MAP of 86\%, accuracy, precision, recall, and F1 score of 79\%, respectively.
This empirical evidence postulates that, within the constraints of our dataset, the data augmentation applied might not be as pivotal in learning accident-associated features.

\subsection{I3D-CONVLSTM2D RGB + Optical-Flow (Trainable)}
Transfer learning is a core machine learning technique that extends knowledge acquired from a source domain to a related but distinct target domain, often yielding improved accuracy \cite{saravanarajan2023car}. In our study, the I3D-CONVLSTM2D RGB + Optical-Flow (Trainable) model effectively employed transfer learning by incorporating the I3D model as a foundational component, with the N-15 layers set to trainable~\cite{Carreira2017}. This strategic approach enabled us to harness the pre-existing knowledge within the I3D model, significantly augmenting our accident detection system proficiency in extracting meaningful features from both RGB and Optical Flow data.
We implemented several crucial enhancements, including batch normalization layers before Global Average Pooling, which expedited the model's convergence during training. Additionally, we increased the number of neurons in dense layers to 512, 256, and 256, respectively, optimizing feature extraction and representation. Notably, we introduced additional filters in the first ConvLSTM2D layer, which played a pivotal role in improving the model's performance.
These enhancements resulted in a substantial leap in performance, with the model achieving an impressive mean average precision (MAP) of 87\%, accuracy, precision, recall, and F1 score of 80\% respectively, as shown in Table (\ref{Results}) reflecting the model's capability in detecting accidents in a video stream.

\subsection{DenseNet-Transformer RGB Only}
In benchmarking our model against other accident detection architectures, we integrated DenseNet121~\cite{huang2017densely}, renowned for robust feature extraction capabilities, accentuated by its densely connected layers. By adjusting videos to 224x224 dimensions, we harmonized computational efficiency with feature richness. These extracted features then flowed into a Transformer architecture~\cite{huang2017densely}, leveraging its self-attention mechanism, allowing targeted emphasis on specific video features crucial for accident detection. The architecture trained on our dataset implemented dense-net for feature extraction, transformer layers, dense layers, and a fully connected classification layer. The model had a precision, recall, and F1 score of 75\% within a longer training timeframe, as shown in Table (\ref{Results}). This configuration underscores the computational intricacies of the real-time feature extraction process.



\section{Discussion and Conclusion}
Accident detection methods have evolved significantly over time. While traditional approaches, such as human-based reporting, remain relevant, there is an unmistakable shift towards modern automated systems. These state-of-the-art systems harness the power of sensors, machine learning algorithms, and computer vision. With its intrinsic capacity to detect accidents in real time and adapt to variable road scenarios, the latter stands out as an innovative solution in the dynamic landscape of accident detection. With relentless technological advancement, there is little doubt that these systems will become indispensable in enhancing traffic safety.
Our exploration culminated with the I3D-CONVLSTM2D Trainable RGB + Optical Flow model, which demonstrated superior performance metrics. Boasting an accuracy of 0.80 and a mean average precision of 87\%, this model exhibited a profound ability to isolate traffic accident features in traffic scenarios interspersed with normal traffic and accident-related events.
Traffic accidents remain a significant concern, especially in densely populated regions where they account for a substantial proportion of fatalities. To address this issue, this study sought to devise a vision-based accident detection system tailored for real-time deployment on edge IoT devices, such as Raspberry Pi. Recognizing the intrinsic challenges of such an approach, notably the massive data requirement, we took the initiative to curate a novel accident dataset. This resource can either complement existing datasets or be employed as a standalone tool, thereby granting researchers the flexibility to extend or modify our foundational framework for accident detection.
While the I3D two-stream network, trained on the Kinetics dataset with 25 million parameters and its extensive training process across 32 GPUs for 110k steps, is computationally demanding. In contrast, our model, designed to be efficient and resource-conscious, was trained on an Ubuntu 20.04.2 LTS system leveraging 2 GPUs, each of 11 GB. The model specifications underscore our commitment to efficiency: the RGB model consists of 3 million parameters, and the I3D-CONVLSTM2D Trainable RGB + Optical extends to 9 million parameters.
In essence, our research has successfully bridged the gap between computational power and real-world applicability. By offering a cost-effective and reliable accident detection system that can be deployed in real-time within a smart city framework, we pave the way for more accessible and ubiquitous surveillance solutions. Most notably, our model simplicity, efficiency, and reduced computational demands, especially when juxtaposed against heavyweights like I3D, serve as a testament to our objective of creating lightweight yet effective solutions for critical societal challenges.

\section{Limitations and Future Work}
Our evaluation showcased that stationary vehicles, whether parked or halted along the road, obscure and sometimes interfere with the visibility of an accident scene, potentially causing the detection system to flag such video input as a traffic accident. Differentiating between regular slow-moving traffic and traffic slowed due to an accident presents another challenge, as patterns in the former can often resemble post-accident scenes. Environmental factors, like dust, sand, smoke, or wind, often obscure vital details and make it difficult for the detection system to clearly identify and classify incidents.
Considering these limitations, future work would focus on refining the accident detection systems. This includes implementing advanced algorithms that can better discern between stationary obstacles and genuine accidents, as well as traffic patterns. Additionally, building algorithms robust enough to adjust for environmental interference will ensure that the scene remains interpretable, regardless of external conditions. These improvements aim to bolster the reliability and precision of traffic accident detection systems in a myriad of real-world scenarios.


\bibliography{main}\bibliographystyle{ieeetr}

\end{document}